\documentclass[runningheads]{llncs}
\usepackage[T1]{fontenc}
\usepackage{graphicx,verbatim}

\usepackage{amsmath}
\usepackage{amssymb}
\usepackage{mathtools}
\usepackage{multirow}
\usepackage{booktabs}

\usepackage{array}
\newcolumntype{L}[1]{>{\raggedright\arraybackslash}p{#1}}
\newcolumntype{C}[1]{>{\centering\arraybackslash}p{#1}}
\begin{document}
\title{Which Tool Response Should I Trust? Tool-Expertise-Aware Chest X-ray Agent with Multimodal Agentic Learning}
%

\author{Zheang Huai, Honglong Yang, and Xiaomeng Li}  
\authorrunning{Z. Huai et al.}
\institute{The Hong Kong University of Science and Technology, Kowloon, Hong Kong \\
    }
  
\maketitle              
\begin{abstract}
AI agents with tool-use capabilities show promise for integrating the domain expertise of various tools. In the medical field, however, tools are usually AI models that are inherently error‑prone and can produce contradictory responses. Existing research on medical agents lacks sufficient understanding of the tools’ realistic reliability and thus cannot effectively resolve tool conflicts. To address this gap, this paper introduces a framework that enables an agent to interact with tools and empirically learn their practical trustworthiness across different types of multimodal queries via agentic learning. As a concrete instantiation, we focus on chest X-ray analysis and present a tool-expertise-aware chest X-ray agent (TEA-CXA). When tool outputs disagree, the agent experimentally accepts or rejects multimodal tool results, receives rewards, and learns which tool to trust for each query type. Importantly, TEA-CXA extends existing codebases for reinforcement learning with multi-turn tool-calling that focus on textual inputs, to support multimodal contexts effectively. In addition, we enhance the codebase for medical use scenarios by supporting multiple tool calls in one turn, parallel tool inference, and multi-image accommodation within a single user query. Our code framework is applicable to general medical research on multi-turn tool-calling reinforcement learning in multimodal settings. Experiments show that TEA-CXA outperforms the state-of-the-art methods and a comprehensive set of baselines. Code will be released.

\end{abstract}
\section{Introduction}
\label{sec:intro}

Large language models (LLMs) and Multi-Modal Large Language Models (MLLMs) have demonstrated remarkable capabilities in decision-making in complex interactive environments~\cite{agentbench}. On the other hand, task-specific models have shown success in automating
various aspects of medical interpretation~\cite{tanno2025collaboration,cohen2022torchxrayvision,rajpurkar2017chexnet}. However, the isolated nature of these models has limited their widespread application in real-world clinical settings.


Medical AI agents with tool-use have emerged as a solution to integrate the domain expertise of various AI models. An agent is driven by a policy model, typically an LLM or MLLM. It can understand user queries, make decisions, and choose appropriate invokable models, hereafter
referred to as ``tools'', to obtain expert opinions or carry out specific tasks~\cite{mmedagent,txagent,xie2024large,chen2023llava}.

\begin{figure}[t]
  \centering
   \includegraphics[width=0.95\textwidth]{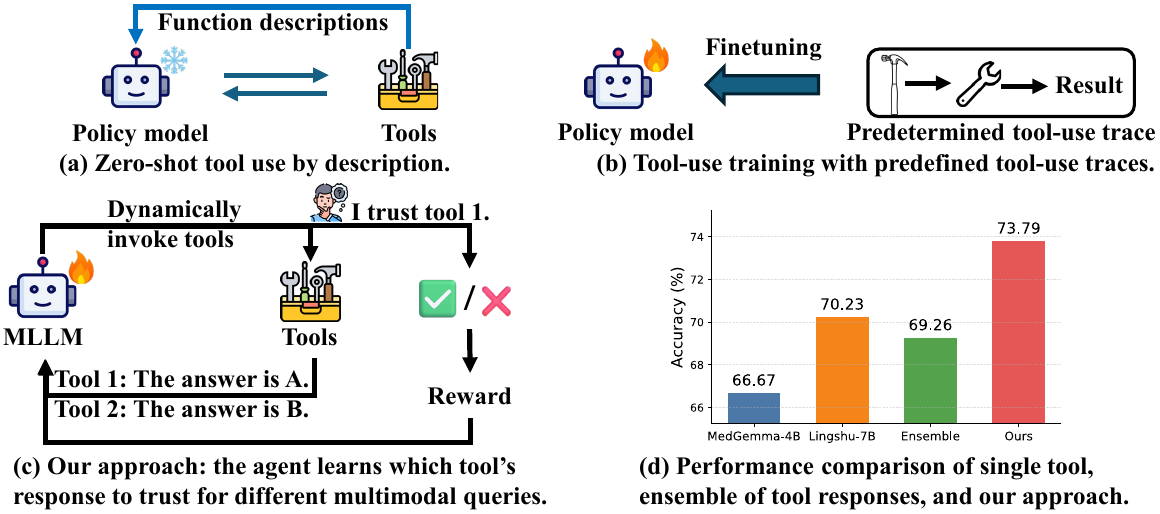}

   \caption{(a)(b) Previous works use tools in a zero-shot manner or fine-tune the policy model with pre-specified tool-use traces, and thus lack sufficient understanding of the tools' realistic reliability and cannot effectively resolve tool conflicts. (c) Our approach enables the agent to learn tools' empirical trustworthiness across different queries through multimodal agentic learning. (d) Our method outperforms any individual tool and agent-based ensembling of tool outputs.}
   \label{fig:intro}
\end{figure}

Current research on medical agents with tool-use primarily falls into two categories, as illustrated in Fig.~\ref{fig:intro}(a)(b). The first category equips a policy model with a suite of external tools and enables it to invoke these tools and integrate their outputs in a zero-shot manner based solely on their functional descriptions. For example, \cite{medrax,agentmd,medorch,zhao2025agentic,ferber2025development} curate domain-specific expert tools tailored to particular tasks and design agents with a frozen policy model. The second category focuses on fine-tuning the policy model with pre-constructed tool-calling and tool output integration traces. For instance, \cite{mmedagent,cpathagent,pathfinder,txagent,voxelprompt} construct datasets with complete tool-use traces and fine-tune the policy model to enable it to invoke suitable tools with appropriate arguments and incorporate tool responses.


However, prior research on medical agents lacks sufficient understanding of the tools’ real-world reliability. The tools commonly used in medical settings are AI models that are inherently prone to inaccuracies~\cite{healthbench}, and may yield contradictory outputs. Their performance also varies across datasets~\cite{medagentsbench}. Relying solely on tool descriptions or pre-constructed tool-use traces prevents agents from having access to tools' practical trustworthiness on the target dataset, leaving them unable to determine which output is correct when conflicts arise. This underscores the need for medical agents to develop a nuanced understanding of each tool’s real-world strengths and limitations across different types of multimodal queries, enabling them to dynamically and appropriately trust the tool responses.

In this paper, we introduce an agentic learning paradigm that enables a multimodal medical agent to empirically learn tools' practical reliability across various types of queries through active interaction with them, thereby learning to trust the correct tool when their outputs conflict. As a concrete instantiation, we study visual question answering (VQA) for chest X-rays (CXRs) and develop a tool-expertise-aware chest X-ray agent (TEA-CXA). Our framework employs reinforcement learning (RL)~\cite{kaelbling1996reinforcement,sutton1999reinforcement} to allow the MLLM to experimentally accept or reject multimodal tool outputs when they disagree, obtain reward scores, and learn which tool to trust for different types of multimodal queries, as shown in Fig.~\ref{fig:intro}(c). As shown in Fig.~\ref{fig:intro}(d), after training, TEA-CXA acquires awareness of each tool's real-world expertise, and can select the appropriate tool response to trust, surpassing the performance of any single tool\footnote{Note that our method is designed for general medical agents where multiple tools with diverse functionalities are \textit{dynamically} (e.g., varying tool types and arguments) invoked. In our experiments, we use two VQA tools to make the RL training affordable (since training time increases markedly with the number of tools) and to facilitate clear evaluation of tool-response selection performance.}. Existing general-purpose tool-using RL frameworks are designed for text-only inputs~\cite{rlfactory,verl}. We extend and refine these frameworks to support the interaction between MLLM and multimodal tools. Furthermore, our enhanced codebase supports multiple tool calls per turn, parallel tool inference, and image selection for tool invocation when multiple images are in a single user query, which are capabilities well-tailored to medical scenarios. The resulting code framework is flexible and applicable to general scenarios involving RL for multi-turn tool-calling in multimodal contexts. Our algorithm does not rely on any rollout data for supervision. It uses only a simple rule-based reward function on the final answer to guide optimization.   

Our contributions are summarized as follows. (1) Pioneering the consideration of tools' real-world trustworthiness to resolve conflicts among tool responses, moving beyond reliance solely on tools' functional descriptions or pre-specified tool-use traces. (2) Proposing to enable the agent to empirically learn tools' practical reliability across different query types via multimodal agentic learning. (3) Designing a robust code framework for multimodal agentic learning, which is applicable to general multi-turn tool-calling RL in multimodal contexts. (4) Evaluation of TEA-CXA on chest X-ray VQA datasets, demonstrating the superiority of our approach over state-of-the-art methods and a variety of baselines.

\section{Method}
Fig.~\ref{fig:method} illustrates our tool-expertise-aware chest X-ray agent (TEA-CXA) framework via multimodal agentic learning. 
In Sec.~\ref{method_sec1}, we introduce tool-expertise-awareness training, which informs the agent about tools' real-world trustworthiness across different types of multimodal queries.
In Sec.~\ref{method_sec2}, we present our agentic learning codebase design that allows an MLLM to interact with multimodal tools and incorporates several practically useful enhancements tailored to medical scenarios.


\subsection{Tool-expertise-awareness training via multimodal agentic learning with multi-turn tool calling}\label{method_sec1}
Tools for medical image analysis are inherently error-prone and can produce contradictory outputs. Relying solely on functional descriptions of tools~\cite{medrax} or pre-specified tool-use traces fails to adequately inform an MLLM about the tools' real-world performance and reliability, and thus does not help it decide which tool output to trust when their responses conflict. To bridge this gap, we propose tool-expertise-awareness training, where we leverage reinforcement learning (RL) to enable the agent to actively interact with tools to learn their real-world trustworthiness for various types of multimodal queries.

\noindent\textbf{Reinforcement learning.}
We adopt Group Relative Policy Optimization (GRPO)~\cite{grpo} as the RL algorithm. Specifically, for each input prompt $x$, GRPO samples a group of trajectories $\{\tau_1, \tau_2, \ldots, \tau_G\}$ from the reference policy $\pi_{\mathrm{ref}}$. Then the advantage within the group $\hat{A}_{i,t}$ is computed as:
\begin{equation}
\hat{A}_{i,t} = \frac{R_\phi(\tau) - \mathrm{mean}(\{R_\phi(\tau_1), \ldots, R_\phi(\tau_G)\})}{\mathrm{std}(\{R_\phi(\tau_1), \ldots, R_\phi(\tau_G)\})},
\end{equation}
where $R_\phi$ is the reward function. The policy model is then optimized by maximizing the following objective function:
\begin{equation}
     J(\theta) \!=\! \frac{1}{G} \!\sum_{i=1}^{G} \!\frac{1}{|\tau_i|}\! 
    \sum_{t=1}^{|\tau_i|} \!\min \!\big[ r_{i,t}(\theta) \hat{A}_{i,t},
    \mathrm{clip}(r_{i,t}(\theta), 1\!-\!\epsilon, 1\!+\!\epsilon) \hat{A}_{i,t} \big] \!-\! \beta \mathrm{D}_{\mathrm{KL}} \left( \pi_\theta \!\parallel\! \pi_{\mathrm{ref}} \right),
\end{equation}
where $\pi_\theta$ is the policy MLLM, $\mathbb{D}_{\mathrm{KL}}$ is the KL divergence, and $r_{i,t}$ is the policy ratio
$r_{i,t}(\theta) = \frac{\pi_\theta(\tau_{i,(t)} \mid \tau_{i,<t})}{\pi_{\mathrm{old}}(\tau_{i,(t)} \mid \tau_{i,<t})}$.

\begin{figure}[t]
  \centering
   \includegraphics[width=\textwidth]{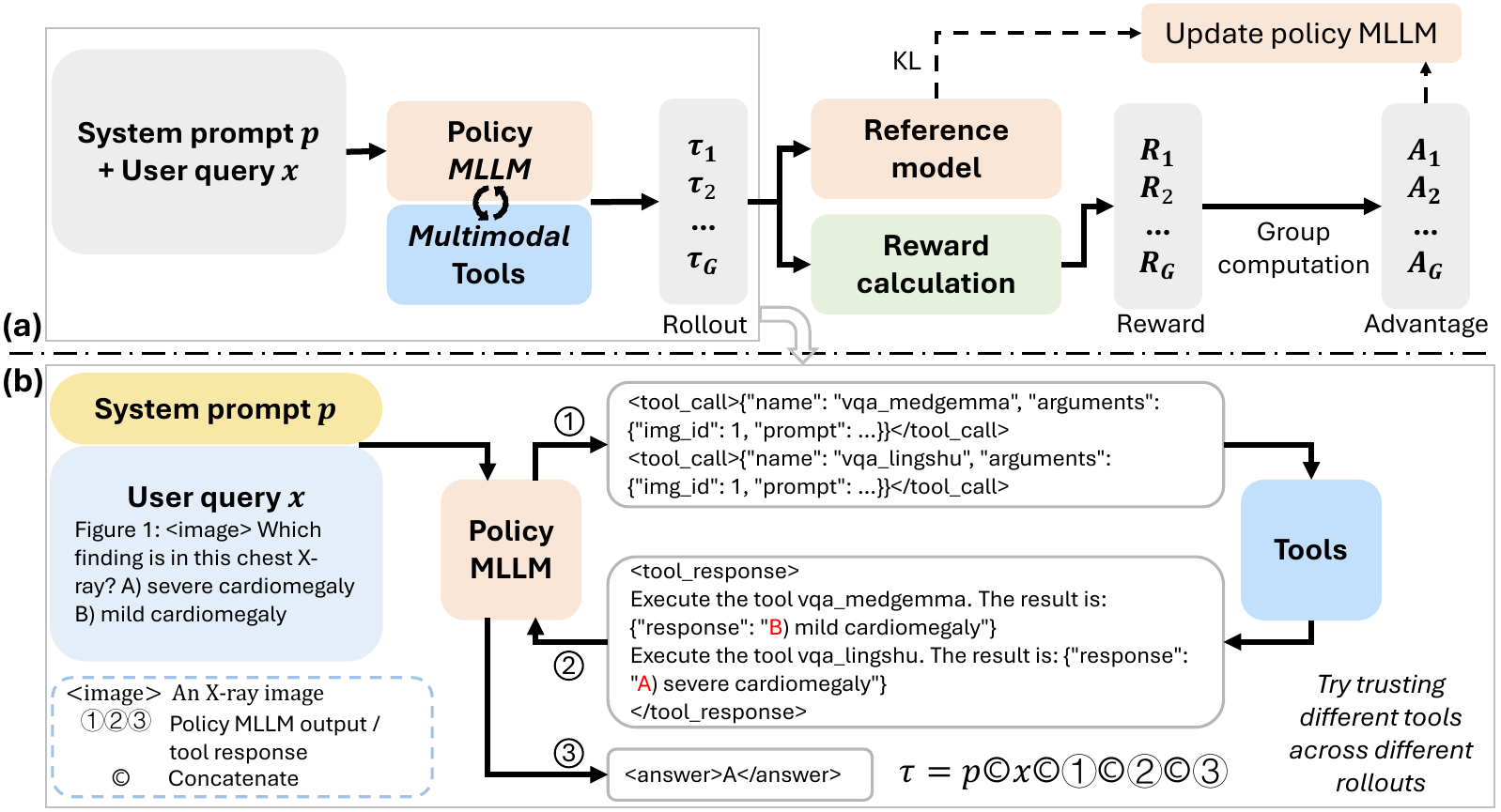}

   \caption{Overview of the proposed tool-expertise-aware chest X-ray agent (TEA-CXA) framework. (a) The agentic learning pipeline with multimodal policy model and multimodal tools. (b) Details of the generation process for a \textit{single} rollout. Tool invocations are dynamically generated by MLLM. Different rollouts try out trusting different tools when tool results contradict.}
   \label{fig:method}
\end{figure}

\noindent\textbf{Tool-expertise-awareness enhancement via multimodal agentic learning with tool calling.}
We aim to boost the MLLM's awareness of tools' practical reliability across different types of multimodal queries, and enable it to dynamically and correctly select the tool response to trust when tool results contradict. Our framework lets the agent call multiple tools, receive their responses, experimentally trust one of the tool outputs if they disagree, and receive reward scores. In this way, the agent obtains a sense of the tools' realistic accuracies for different types of user queries and thus can select appropriate tools to trust.

As illustrated in Fig.~\ref{fig:method}, the rollout $\tau$ in this paper is the concatenation of interleaved MLLM-generated tool-calling tokens and tool responses. We use \texttt{<tool\_call>}
and \texttt{</tool\_call>} to enclose a tool-calling action that is written as a JSON object, which will be parsed and executed by the tools, with the tool execution results enclosed in \texttt{<tool\_response>} and \texttt{</tool\_response>}. Then, the existing rollout concatenated with the tool responses will be used as the input to generate the MLLM response in the next turn. The rollout process stops when the MLLM outputs the final answer, which is wrapped in \texttt{<answer></answer>}. 

\begin{figure}[t]
  \centering
   \includegraphics[width=\textwidth]{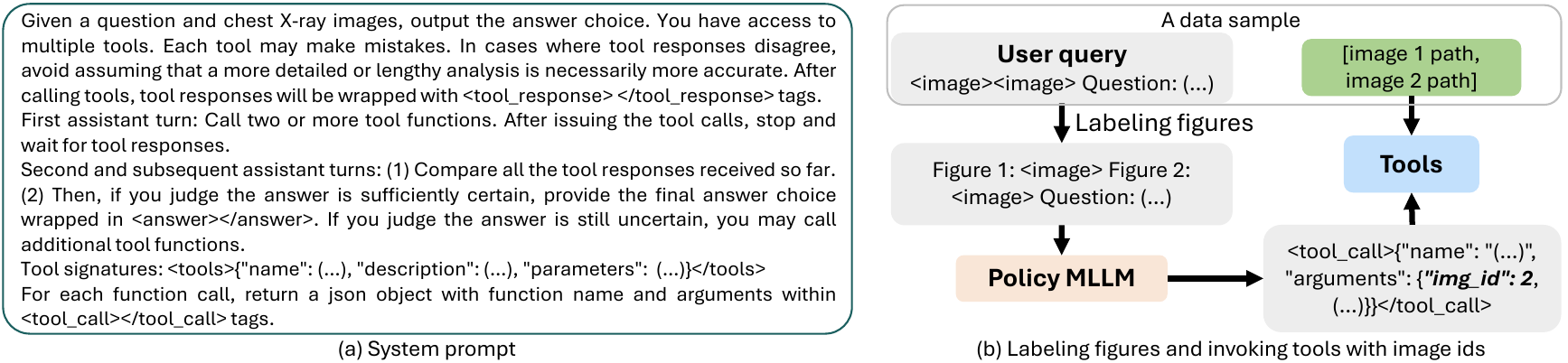}

   \caption{(a) The system prompt instructs the policy model to follow the specified output format and consider each tool’s output as potentially trustworthy when tool results contradict. (b) For queries containing multiple images, our approach directs the policy model to generate image indices instead of file paths for efficient tool invocation. Some details are abbreviated as ``(...)'' due to space constraints.}
   \label{fig:prompt}
\end{figure}

We design a system prompt that guides the policy MLLM to produce rollouts in the defined format, as demonstrated in Fig.~\ref{fig:prompt}(a). In this prompt, the MLLM is instructed to experimentally and randomly trust one tool if tool responses conflict. This prevents the MLLM from biasing toward a specific tool (e.g., one that provides more detailed analysis), given that each tool may have its strengths and weaknesses on different types of queries.

We use a tool-calling format reward $R_t(\tau)$ to verify whether the MLLM adheres to our prescribed tool-calling format, and an answer format reward $R_a(\tau)$ to check the existence of the \texttt{<answer></answer>} tags in the rollout. $R_t(\tau)$ and $R_a(\tau)$ are set to 0.1 if their respective conditions are satisfied, and 0 otherwise. The total reward is the sum of the outcome reward and the two format rewards:
\begin{equation}
    R(\tau) = R_o(\tau)+R_a(\tau)+R_t(\tau),
\end{equation}
where the outcome reward $R_o(\tau)$ evaluates the final answer using exact matching. $R_o(\tau)$ is 1 if the answer is correct, and 0 otherwise.

The rollout sequence consists of both the MLLM-generated tokens and tool responses. We apply loss masking for the tool responses~\cite{searchr1}, ensuring the policy objective gradient is computed only over MLLM-generated tokens.

\subsection{Multimodal agentic learning framework design}\label{method_sec2}
We extend existing agentic training codebases focused on text~\cite{rlfactory,verl} to robustly support multimodal contexts, with enhancements tailored for medical scenarios. The framework is extensible to other domains beyond X-ray analysis, providing a solid base for future research on medical tool-use agents.

\noindent\textbf{Interaction between MLLM and multimodal tools in RL.}
Existing code frameworks for reinforcement learning with tool calling, e.g., RL-Factory~\cite{rlfactory}, center on LLM policy models and text-only tools, lacking verifiable multimodal support. Building on RL-Factory~\cite{rlfactory}, we expand and refine functionality to enable reliable interaction between MLLM policy model and multimodal tools. Our code will be released.

\noindent\textbf{Multiple tool calls per turn and parallel tool inference.}
Given the imperfect accuracy of medical AI tools, it is beneficial to consult more than one tool in a single turn before proceeding to the next step. We extend \cite{rlfactory} to allow for multiple tool calls per turn, each enclosed in \texttt{<tool\_call>} and
\texttt{</tool\_call>}. Tool results are concatenated and returned to the MLLM. In addition, some medical AI tools have large numbers of parameters and slow inference, becoming the training time bottleneck. To speed up tool responses and accelerate training, our code framework deploys multiple instances of server APIs of the same tool on different ports and performs port round-robin when making calls.

\noindent\textbf{Accommodation for multi-image queries.}
Although we validate our method on single-image queries, our code framework is readily applicable to multi-image queries. A single query may contain more than one image (e.g., CXRs from different views, including AP, PA, Lateral), while some medical tools accept only one or few images~\cite{medgemma}. To allow for image selection when calling tools, we label images in the chat template (e.g., ``Figure 1'', ``Figure 2'') and expose the image label as a tool argument. All image paths of a data sample are passed implicitly as a hidden argument. Fig.~\ref{fig:prompt}(b) illustrates this workflow. The benefits are two-fold. First, MLLMs can reliably produce short labels, avoiding errors while generating long path strings. Second, this reduces token overhead. The multi-image functionality is verified on ReXVQA~\cite{rexvqa}, though quantitative results are not included due to space constraints.

\begin{table}[!t]
  \caption{Accuracies (\%) on CheXbench, which consists of three subsets: Rad-Restruct, SLAKE, and OpenI. ``MedRAX'' denotes its original implementation; ``MedRAX\textsuperscript{*}'' denotes using the method in MedRAX but applying the same policy MLLM and tools as ours. Note that MedRAX reports mean per-class accuracy, while we use overall accuracy as metric. }
  \label{tab:chexbench}
  \centering
  \begin{tabular}{C{4.2cm}|C{2.0cm}|C{2.0cm}C{1.5cm}C{1.6cm}}
    \toprule
    Method & \textbf{Overall} & Rad-Restruct & SLAKE & OpenI \\
    \hline
    Qwen2.5-VL-7B-Instruct~\cite{qwen2vl}& 52.4 & 55.7 & 70.7 & 45.5 \\
    MedGemma-4B~\cite{medgemma}& 66.7 & 60.0 & 78.9 & 64.7 \\
    Lingshu-7B~\cite{lingshu}& 70.2 & 65.2 & 92.7 & 64.5 \\
    Agent-ensemble& 69.3 & 67.0 & 87.8 & 63.9 \\
    Reasoning~\cite{deepseekr1}& 62.8 & 65.2 & 91.9 & 52.6 \\
    Ensemble-A & 67.6 & 60.9 & 88.6 & 62.9 \\
    Ensemble-B& 69.7 & 61.7 & 94.3 & 64.2 \\
    CheXagent~\cite{chexagent} & 62.4 & 57.1 & 78.1 & 59.0 \\
    GPT-4o & 58.4 & 53.9 & 85.4 & 51.1 \\
    MedRAX~\cite{medrax} & 61.6 & 68.7 & 82.9 & 52.6 \\
    MedRAX\textsuperscript{*}& 69.6 & 61.7 & 90.2 & 65.3 \\
    TEA-CXA (Ours) & \textbf{73.8} & \textbf{69.6} & \textbf{95.9} & \textbf{67.9} \\
    \bottomrule
  \end{tabular}
\end{table}

\begin{figure}[t]
  \centering
   \includegraphics[width=\textwidth]{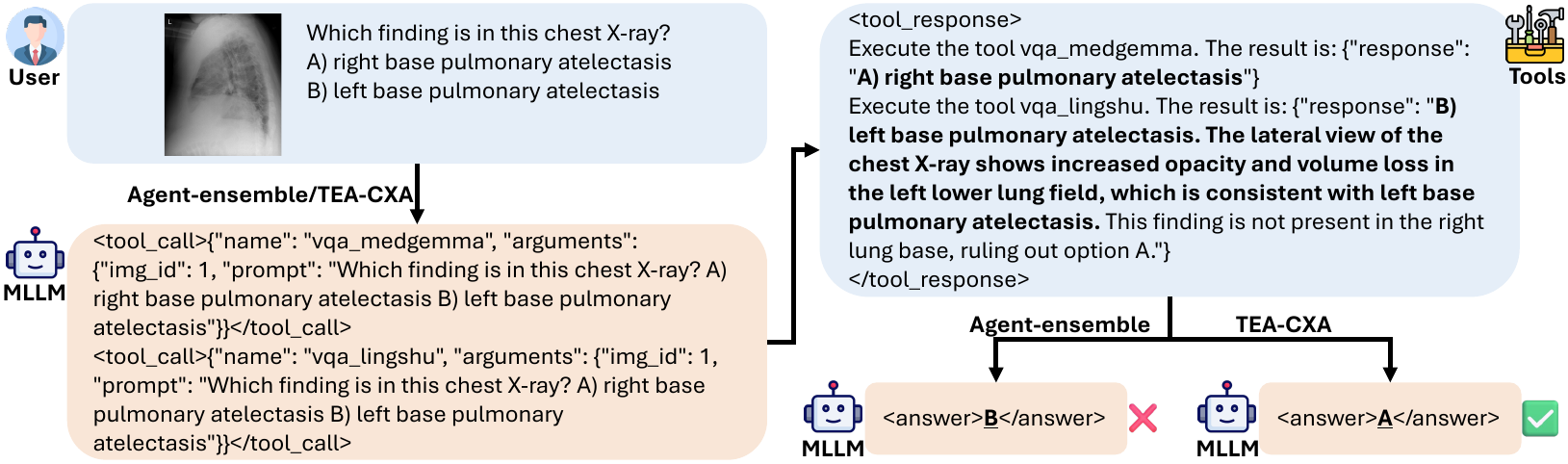}

   \caption{Qualitative comparison on a sample in CheXbench. Although Lingshu offers more detailed justifications, our method correctly trusts MedGemma’s answer, thanks to its awareness of tools' realistic reliability across multimodal query types.}
   \label{fig:quali}
\end{figure}

\section{Experiments}
\noindent\textbf{Datasets.}
We evaluate TEA-CXA on CheXbench~\cite{chexagent} for the visual question answering (VQA) task. We follow MedRAX~\cite{medrax} to use the three subsets of CheXbench: Rad-Restruct, SLAKE, and OpenI, comprising a total of 618 multiple-choice questions. We conduct three-fold cross-validation experiments on CheXbench and ensemble the results across the three folds. Model performance is measured by accuracy (percentage of correct answers).

\noindent\textbf{Implementations.}
TEA-CXA employs Qwen2.5-VL-7B-Instruct~\cite{qwen2vl} as its policy MLLM. We equip the agent with two high-performance VQA tools: MedGemma-4B~\cite{medgemma} and Lingshu-7B~\cite{lingshu}, to assess the performance gain of our approach with up-to-date tools. Each tool is implemented according to its official HuggingFace guidelines. Training is conducted on 8 NVIDIA H800 GPUs. We train TEA-CXA for 100 steps with a batch size of 80. More details are available in our code.

\noindent\textbf{Quantitative performance.}
We compare our method against the following baselines and models: (1) Direct inference on Qwen2.5-VL-7B-Instruct~\cite{qwen2vl} (our policy MLLM), MedGemma~\cite{medgemma}, or Lingshu~\cite{lingshu} (our tools); (2) Agent-ensemble: invoking both tools and ensembling their outputs with the same policy MLLM, which acts as a naïve agent strategy; (3) Reasoning: training the same policy MLLM to reason and provide the answer with RL~\cite{deepseekr1}. (4) Policy MLLM-tool ensembles: Ensemble-A, which combines the results of Qwen2.5-VL-7B-Instruct and tools, and Ensemble-B, which combines the RL-trained reasoning policy MLLM from (3) with tool outputs. (5) State-of-the-art (SOTA) methods, including biomedical vision-language model CheXagent~\cite{chexagent}, general-purpose MLLM GPT-4o, and CXR-focused agent MedRAX~\cite{medrax}. The results are shown in Table~\ref{tab:chexbench}, TEA-CXA outperforms all baselines and prior SOTA.

\noindent\textbf{Qualitative performance.}
We present one representative case  in Fig.~\ref{fig:quali} comparing TEA-CXA to Agent-ensemble (invoking both tools and ensembling their outputs). The two tools provide inconsistent suggestions. Agent-ensemble incorrectly favors Lingshu, likely due to its more detailed analysis. TEA-CXA correctly trusts MedGemma despite its concise conclusion without analysis. This demonstrates TEA-CXA's ability to resolve conflicting tool outputs based on its awareness of tools' real-world trustworthiness across different multimodal query types instead of surface features of tool outputs.

\begin{table}[!t]
  \caption{Tool response selection accuracies (\%) for samples where tool outputs conflict and at least one output is correct. }
  \label{tab:tool_selection}
  \centering
  \begin{tabular}{C{2.6cm}|C{2.0cm}|C{2.0cm}C{2.0cm}C{1.5cm}}
    \toprule
    Method & \textbf{Overall} & Rad-Restruct & SLAKE & OpenI \\
    \hline
    Ensemble-A & 46.6 & 42.3 & 62.1 & 43.7 \\
    Ensemble-B& 54.6 & 46.2 & 86.2 & 48.7 \\
    Agent-ensemble& 54.0 & 57.7 & 58.6 & 52.1 \\
    MedRAX\textsuperscript{*}& 57.5 & 42.3 & 75.9 & 55.5 \\
    TEA-CXA (Ours) & \textbf{63.8} & \textbf{69.2} & \textbf{89.7} & \textbf{56.3} \\
    \bottomrule
  \end{tabular}
\end{table}

\noindent\textbf{Tool response selection analysis.}
Table~\ref{tab:tool_selection} presents the accuracies of choosing the correct tool response among conflicting outputs when at least one is correct. The compared methods have no prior knowledge about the tools' real-world performance and therefore can only rely on superficial features in the tool outputs to resolve conflicts. In contrast, our approach, trained via multimodal agentic learning to internalize each tool's reliability across different query types, selects the correct tool output with substantially higher accuracy.

\section{Conclusion}
This work presents a novel tool-expertise-aware chest X-ray agent. The agent empirically learns tools' practical trustworthiness via agentic learning. When tool outputs conflict, It experimentally trusts one
tool result, receives rewards, and then learns which tool to trust for each query type. To implement our method, we develop a code framework for multimodal agentic learning with enhancements for medical scenarios, building a foundation for future research.
Experiments show that our method
outperforms the state-of-the-art methods and strong baselines.

\bibliographystyle{splncs04}
\bibliography{mybibliography}

@String(AAAI = {AAAI})

@article{mmedagent,
  title={Mmedagent: Learning to use medical tools with multi-modal agent},
  author={Li, Binxu and Yan, Tiankai and Pan, Yuanting and Luo, Jie and Ji, Ruiyang and Ding, Jiayuan and Xu, Zhe and Liu, Shilong and Dong, Haoyu and Lin, Zihao and others},
  journal={arXiv preprint arXiv:2407.02483},
  year={2024}
}

@article{voxelprompt,
  title={VoxelPrompt: A Vision-Language Agent for Grounded Medical Image Analysis. Oct. 10, 2024, arXiv},
  author={Hoopes, A and Butoi, VI and Guttag, JV and Dalca, AV},
  journal={arXiv preprint arXiv:2410.08397}
}

@article{medrax,
  title={MedRAX: Medical Reasoning Agent for Chest X-ray},
  author={Fallahpour, Adibvafa and Ma, Jun and Munim, Alif and Lyu, Hongwei and Wang, Bo},
  journal={arXiv preprint arXiv:2502.02673},
  year={2025}
}

@article{grpo,
  title={Deepseekmath: Pushing the limits of mathematical reasoning in open language models},
  author={Shao, Zhihong and Wang, Peiyi and Zhu, Qihao and Xu, Runxin and Song, Junxiao and Bi, Xiao and Zhang, Haowei and Zhang, Mingchuan and Li, YK and Wu, Yang and others},
  journal={arXiv preprint arXiv:2402.03300},
  year={2024}
}

@article{searchr1,
  title={Search-r1: Training llms to reason and leverage search engines with reinforcement learning},
  author={Jin, Bowen and Zeng, Hansi and Yue, Zhenrui and Yoon, Jinsung and Arik, Sercan and Wang, Dong and Zamani, Hamed and Han, Jiawei},
  journal={arXiv preprint arXiv:2503.09516},
  year={2025}
}

@article{rlfactory,
  title={Rlfactory: A plug-and-play reinforcement learning post-training framework for llm multi-turn tool-use},
  author={Chai, Jiajun and Yin, Guojun and Xu, Zekun and Yue, Chuhuai and Jia, Yi and Xia, Siyu and Wang, Xiaohan and Jiang, Jiwen and Li, Xiaoguang and Dong, Chengqi and others},
  journal={arXiv preprint arXiv:2509.06980},
  year={2025}
}

@article{verl,
  title   = {HybridFlow: A Flexible and Efficient RLHF Framework},
  author  = {Guangming Sheng and Chi Zhang and Zilingfeng Ye and Xibin Wu and Wang Zhang and Ru Zhang and Yanghua Peng and Haibin Lin and Chuan Wu},
  year    = {2024},
  journal = {arXiv preprint arXiv: 2409.19256}
}

@article{tanno2025collaboration,
  title={Collaboration between clinicians and vision--language models in radiology report generation},
  author={Tanno, Ryutaro and Barrett, David GT and Sellergren, Andrew and Ghaisas, Sumedh and Dathathri, Sumanth and See, Abigail and Welbl, Johannes and Lau, Charles and Tu, Tao and Azizi, Shekoofeh and others},
  journal={Nature Medicine},
  volume={31},
  number={2},
  pages={599--608},
  year={2025},
  publisher={Nature Publishing Group US New York}
}

@article{rajpurkar2017chexnet,
  title={Chexnet: Radiologist-level pneumonia detection on chest x-rays with deep learning. arXiv},
  author={Rajpurkar, Pranav and Irvin, Jeremy and Zhu, Kaylie and Yang, Brandon and Mehta, Hershel and Duan, Tony and Ding, Daisy and Bagul, Aarti and Langlotz, Curtis and Shpanskaya, Katie and others},
  journal={arXiv preprint arXiv:1711.05225},
  volume={10},
  year={2017}
}

@inproceedings{cohen2022torchxrayvision,
  title={TorchXRayVision: A library of chest X-ray datasets and models},
  author={Cohen, Joseph Paul and Viviano, Joseph D and Bertin, Paul and Morrison, Paul and Torabian, Parsa and Guarrera, Matteo and Lungren, Matthew P and Chaudhari, Akshay and Brooks, Rupert and Hashir, Mohammad and others},
  booktitle={International Conference on Medical Imaging with Deep Learning},
  pages={231--249},
  year={2022},
  organization={PMLR}
}

@article{xie2024large,
  title={Large multimodal agents: A survey},
  author={Xie, Junlin and Chen, Zhihong and Zhang, Ruifei and Wan, Xiang and Li, Guanbin},
  journal={arXiv preprint arXiv:2402.15116},
  year={2024}
}

@article{chen2023llava,
  title={Llava-interactive: An all-in-one demo for image chat, segmentation, generation and editing},
  author={Chen, Wei-Ge and Spiridonova, Irina and Yang, Jianwei and Gao, Jianfeng and Li, Chunyuan},
  journal={arXiv preprint arXiv:2311.00571},
  year={2023}
}

@article{txagent,
  title={TxAgent: An AI agent for therapeutic reasoning across a universe of tools},
  author={Gao, Shanghua and Zhu, Richard and Kong, Zhenglun and Noori, Ayush and Su, Xiaorui and Ginder, Curtis and Tsiligkaridis, Theodoros and Zitnik, Marinka},
  journal={arXiv preprint arXiv:2503.10970},
  year={2025}
}

@article{zhao2025agentic,
  title={An agentic system for rare disease diagnosis with traceable reasoning},
  author={Zhao, Weike and Wu, Chaoyi and Fan, Yanjie and Zhang, Xiaoman and Qiu, Pengcheng and Sun, Yuze and Zhou, Xiao and Wang, Yanfeng and Sun, Xin and Zhang, Ya and others},
  journal={arXiv preprint arXiv:2506.20430},
  year={2025}
}

@article{ferber2025development,
  title={Development and validation of an autonomous artificial intelligence agent for clinical decision-making in oncology},
  author={Ferber, Dyke and El Nahhas, Omar SM and W{\"o}lflein, Georg and Wiest, Isabella C and Clusmann, Jan and Le{\ss}mann, Marie-Elisabeth and Foersch, Sebastian and Lammert, Jacqueline and Tschochohei, Maximilian and J{\"a}ger, Dirk and others},
  journal={Nature cancer},
  pages={1--13},
  year={2025},
  publisher={Nature Publishing Group US New York}
}

@article{sutton1999reinforcement,
  title={Reinforcement learning},
  author={Sutton, Richard S and Barto, Andrew G and others},
  journal={Journal of Cognitive Neuroscience},
  volume={11},
  number={1},
  pages={126--134},
  year={1999}
}

@article{kaelbling1996reinforcement,
  title={Reinforcement learning: A survey},
  author={Kaelbling, Leslie Pack and Littman, Michael L and Moore, Andrew W},
  journal={Journal of artificial intelligence research},
  volume={4},
  pages={237--285},
  year={1996}
}

@article{deepseekr1,
  title={Deepseek-r1: Incentivizing reasoning capability in llms via reinforcement learning},
  author={Guo, Daya and Yang, Dejian and Zhang, Haowei and Song, Junxiao and Zhang, Ruoyu and Xu, Runxin and Zhu, Qihao and Ma, Shirong and Wang, Peiyi and Bi, Xiao and others},
  journal={arXiv preprint arXiv:2501.12948},
  year={2025}
}

@article{qwen2vl,
  title={Qwen2-VL: Enhancing Vision-Language Model's Perception of the World at Any Resolution},
  author={Wang, Peng and Bai, Shuai and Tan, Sinan and Wang, Shijie and Fan, Zhihao and Bai, Jinze and Chen, Keqin and Liu, Xuejing and Wang, Jialin and Ge, Wenbin and Fan, Yang and Dang, Kai and Du, Mengfei and Ren, Xuancheng and Men, Rui and Liu, Dayiheng and Zhou, Chang and Zhou, Jingren and Lin, Junyang},
  journal={arXiv preprint arXiv:2409.12191},
  year={2024}
}

@article{medgemma,
  title={MedGemma Technical Report},
  author={Sellergren, Andrew and Kazemzadeh, Sahar and Jaroensri, Tiam and Kiraly, Atilla and Traverse, Madeleine and Kohlberger, Timo and Xu, Shawn and Jamil, Fayaz and Hughes, Cían and Lau, Charles and others},
  journal={arXiv preprint arXiv:2507.05201},
  year={2025}
}

@article{lingshu,
  title={Lingshu: A Generalist Foundation Model for Unified Multimodal Medical Understanding and Reasoning},
  author={Xu, Weiwen and Chan, Hou Pong and Li, Long and Aljunied, Mahani and Yuan, Ruifeng and Wang, Jianyu and Xiao, Chenghao and Chen, Guizhen and Liu, Chaoqun and Li, Zhaodonghui and others},
  journal={arXiv preprint arXiv:2506.07044},
  year={2025}
}

@article{rexvqa,
  title={ReXVQA: A Large-scale Visual Question Answering Benchmark for Generalist Chest X-ray Understanding},
  author={Pal, Ankit and Lee, Jung Oh and Zhang, Xiaoman and Sankarasubbu, Malaikannan and Roh, Seunghyeon and Kim, Won Jung and Lee, Meesun and Rajpurkar, Pranav},
  journal={arxiv},
  year={2025}
}

@article{agentbench,
  title={Agentbench: Evaluating llms as agents},
  author={Liu, Xiao and Yu, Hao and Zhang, Hanchen and Xu, Yifan and Lei, Xuanyu and Lai, Hanyu and Gu, Yu and Ding, Hangliang and Men, Kaiwen and Yang, Kejuan and others},
  journal={arXiv preprint arXiv:2308.03688},
  year={2023}
}

@article{agentmd,
  title={Agentmd: Empowering language agents for risk prediction with large-scale clinical tool learning},
  author={Jin, Qiao and Wang, Zhizheng and Yang, Yifan and Zhu, Qingqing and Wright, Donald and Huang, Thomas and Khandekar, Nikhil and Wan, Nicholas and Ai, Xuguang and Wilbur, W John and others},
  journal={Nature Communications},
  volume={16},
  number={1},
  pages={9377},
  year={2025},
  publisher={Nature Publishing Group UK London}
}

@article{cpathagent,
  title={CPathAgent: An Agent-based Foundation Model for Interpretable High-Resolution Pathology Image Analysis Mimicking Pathologists' Diagnostic Logic},
  author={Sun, Yuxuan and Si, Yixuan and Zhu, Chenglu and Zhang, Kai and Shui, Zhongyi and Ding, Bowen and Lin, Tao and Yang, Lin},
  journal={arXiv preprint arXiv:2505.20510},
  year={2025}
}

@article{pathfinder,
  title={Pathfinder: A multi-modal multi-agent system for medical diagnostic decision-making applied to histopathology},
  author={Ghezloo, Fatemeh and Seyfioglu, Mehmet Saygin and Soraki, Rustin and Ikezogwo, Wisdom O and Li, Beibin and Vivekanandan, Tejoram and Elmore, Joann G and Krishna, Ranjay and Shapiro, Linda},
  journal={arXiv preprint arXiv:2502.08916},
  year={2025}
}

@article{medorch,
  title={MedOrch: Medical Diagnosis with Tool-Augmented Reasoning Agents for Flexible Extensibility},
  author={He, Yexiao and Li, Ang and Liu, Boyi and Yao, Zhewei and He, Yuxiong},
  journal={arXiv preprint arXiv:2506.00235},
  year={2025}
}

@inproceedings{chexagent,
  title={Chexagent: Towards a foundation model for chest x-ray interpretation},
  author={Chen, Zhihong and Varma, Maya and Delbrouck, Jean-Benoit and Paschali, Magdalini and Blankemeier, Louis and Van Veen, Dave and Valanarasu, Jeya Maria Jose and Youssef, Alaa and Cohen, Joseph Paul and Reis, Eduardo Pontes and others},
  booktitle={AAAI 2024 Spring Symposium on Clinical Foundation Models},
  year={2024}
}

@article{medagentsbench,
  title={Medagentsbench: Benchmarking thinking models and agent frameworks for complex medical reasoning},
  author={Tang, Xiangru and Shao, Daniel and Sohn, Jiwoong and Chen, Jiapeng and Zhang, Jiayi and Xiang, Jinyu and Wu, Fang and Zhao, Yilun and Wu, Chenglin and Shi, Wenqi and others},
  journal={arXiv preprint arXiv:2503.07459},
  year={2025}
}

@article{healthbench,
  title={Healthbench: Evaluating large language models towards improved human health},
  author={Arora, Rahul K and Wei, Jason and Hicks, Rebecca Soskin and Bowman, Preston and Qui{\~n}onero-Candela, Joaquin and Tsimpourlas, Foivos and Sharman, Michael and Shah, Meghan and Vallone, Andrea and Beutel, Alex and others},
  journal={arXiv preprint arXiv:2505.08775},
  year={2025}
}

\end{document}